\documentclass[conference, 10pt]{IEEEtran}
\IEEEoverridecommandlockouts

\usepackage{hhline}
\usepackage[numbers]{natbib}
\usepackage[utf8]{inputenc}
\usepackage{graphicx}
\usepackage{amsfonts}
\usepackage{booktabs}
\usepackage{multirow}
\usepackage{threeparttable}
\usepackage{eurosym,makecell,comment}
\usepackage{url}
\usepackage{xcolor}
\usepackage{tikz}
\usepackage{booktabs}
\usepackage{float}
\usepackage{amsmath}
\usepackage{color, colortbl}
\usepackage{algorithm}
\usepackage{algorithmic}
\usepackage{threeparttable} 

\definecolor{Gray}{gray}{0.9}
\definecolor{LightCyan}{rgb}{0.88,1,1}

\usepackage{array}
\newcolumntype{P}[1]{>{\raggedright\arraybackslash}p{#1}}
\usepackage{dblfloatfix}

\DeclareRobustCommand*{\IEEEauthorrefmark}[1]{%
  \raisebox{0pt}[0pt][0pt]{\textsuperscript{\footnotesize #1}}%
}

\begin{document}

\title{S-VOTE: Similarity-based Voting for Client Selection in Decentralized Federated Learning}


\author{
    \IEEEauthorblockN{Pedro Miguel S\'anchez S\'anchez\IEEEauthorrefmark{1,2,*}, Enrique Tom\'as Mart\'inez Beltr\'an\IEEEauthorrefmark{1}, Chao Feng \IEEEauthorrefmark{3}\\ G\'er\^ome Bovet\IEEEauthorrefmark{4}, Gregorio Mart\'inez P\'erez\IEEEauthorrefmark{1}, Alberto Huertas Celdr\'an\IEEEauthorrefmark{1,3}}  

    \IEEEauthorblockA{\IEEEauthorrefmark{*} Corresponding author.}

    \IEEEauthorblockA{\IEEEauthorrefmark{1}Department of Information and Communications Engineering, University of Murcia, 30100--Murcia, Spain}
    
    \IEEEauthorblockA{\IEEEauthorrefmark{2}Advantx Technological Foundation (Funditec), 28046-Madrid, Spain}

    \IEEEauthorblockA{\IEEEauthorrefmark{3}Communication Systems Group CSG, Department of Informatics, University of Zurich UZH, CH--8050 Zürich, Switzerland}
    
    \IEEEauthorblockA{\IEEEauthorrefmark{4}Cyber-Defence Campus, armasuisse Science \& Technology, CH--3602 Thun, Switzerland}

    \IEEEauthorblockA{[pedromiguel.sanchez, enriquetomas, gregorio]@um.es, gerome.bovet@armasuisse.ch, [cfeng, huertas]@ifi.uzh.ch}
}

\maketitle

\begin{abstract}
Decentralized Federated Learning (DFL) enables collaborative, privacy-preserving model training without relying on a central server. This decentralized approach reduces bottlenecks and eliminates single points of failure, enhancing scalability and resilience. However, DFL also introduces challenges such as suboptimal models with non-IID data distributions, increased communication overhead, and resource usage. Thus, this work proposes S-VOTE, a voting-based client selection mechanism that optimizes resource usage and enhances model performance in federations with non-IID data conditions. S-VOTE considers an adaptive strategy for spontaneous local training that addresses participation imbalance, allowing underutilized clients to contribute without significantly increasing resource costs. Extensive experiments on benchmark datasets demonstrate the S-VOTE effectiveness. More in detail, it achieves lower communication costs by up to 21\%, 4-6\% faster convergence, and improves local performance by 9-17\% compared to baseline methods in some configurations, all while achieving a 14-24\% energy consumption reduction. These results highlight the potential of S-VOTE to address DFL challenges in heterogeneous environments.

\end{abstract}

\begin{IEEEkeywords}
Decentralized Federated Learning, Client Selection, Voting Mechanisms, Communication Efficiency, Model Similarity
\end{IEEEkeywords}

\begin{table*}[h]
\centering
\caption{Comparison of Related Work with the Proposed Approach}
\label{tab:comparison}
\begin{tabular}{p{0.8cm}p{0.7cm}p{4.5cm}p{2.5cm}p{2.5cm}p{4.5cm}}
\hline
\multicolumn{1}{c}{\textbf{Work}} & 
\multicolumn{1}{c}{\textbf{Year}} & 
\multicolumn{1}{c}{\textbf{Approach}} & 
\multicolumn{1}{c}{\textbf{Dataset}} & 
\multicolumn{1}{c}{\textbf{Distribution}} & 
\multicolumn{1}{c}{\textbf{Results}} \\ \hline
\cite{sui2022find} & 2022 & Expectation-maximization method for client utility evaluation and selection & CIFAR10/100, Office-Home & Non-IID Pathological &  $\approx$10-15\% improvement over baseline aggregations and local training\\ \hline

\cite{sadiev2022decentralized} & 2022 & Computational network alignment to reduce resource overhead& LIBSVM Database & IID & Faster model convergence and $\approx$1-2\% baseline improvement \\ \hline

\cite{kharrat2024decentralized} & 2024 & Greedy algorithm-based personalized DFL & CIFACR10, CINIC10, FEMNIST & Non-IID Dir(0.1) and Pathological & $\approx$3-8\% accuracy improvement over local training, baseline FedAvg and other literature methods\\
\hline
\cite{long2024decentralized} & 2024 & Sparse training with dynamic aggregation for efficient personalized DFL & CIFAR10/100, HAM10000 & Non-IID Dir (0.3, 0.5) and Pathological & $\approx$10\% accuracy improvement over literature aggregation methods.  x5 reduction on computation time \\
\hline
\cite{wang2024smart} & 2024 & Neighbor selection based on coresets in DFL & FEMNIST, CIFAR10/100 & Non-IID Dir (0.1, 0.5) and Pathological & $\approx$5-10\% improvement over other literature methods such as \cite{sui2022find}\\
\hline
This work      & 2025          & Similarity and voting-based client selection with adaptive training   & MNIST, FashionMNIST, EMNIST, CIFAR10     & Non-IID Dir(0.1, 0.5) & 9-17\% accuracy improvement, 11-21\% reduction in communication costs, and 14-24\% lower energy consumption \\ \hline
\end{tabular}
\end{table*}

\section{Introduction}

In an era of increasing data privacy concerns, traditional centralized machine learning approaches pose significant risks by requiring raw data to be collected and stored in centralized servers, making it vulnerable to breaches and misuse. In this context, Federated learning (FL) has emerged as a promising solution, enabling collaborative model training across distributed clients without sharing raw data \cite{li2020review}. However, most FL systems rely on a central server for coordination, which introduces bottlenecks, single points of failure, and trust issues \cite{kairouz2021advances}. Decentralized Federated Learning (DFL) addresses these limitations by adopting a peer-to-peer approach, eliminating the need for a central server \cite{beltran2023decentralized}. By leveraging direct client communication, DFL reduces reliance on centralized infrastructures, enhancing resilience, scalability, and privacy.

Despite its advantages, DFL introduces several challenges that hinder its efficiency. More in detail, the two main concerns studied in this paper are i)  the impact of non-IID data, leading to biased or suboptimal global models \cite{hsieh2020non}, and ii) the increased communication overhead and resource usage \cite{liu2022decentralized}. However, these are not the only ones since the absence of a central authority makes it harder to detect and isolate malicious clients manipulating updates to degrade performance or compromise privacy \cite{feng2024dart}. Therefore, addressing these issues is essential to improve resource efficiency, handle data heterogeneity, and ensure trust in decentralized systems.

Several strategies have been proposed to mitigate the challenges posed by DFL. From the communication overhead perspective, techniques such as model quantization and distillation are commonly used to enable a more efficient exchange of information between clients \cite{sanchez2024profe}. From the non-IID data distribution angle, partial aggregation based on model similarity has also been explored. Among these approaches, client selection mechanisms have shown significant promise \cite{fu2023client}. They allow clients to selectively participate in training rounds based on relevance, reducing resource usage and improving convergence. However, most solutions focus on client selection from the server perspective in centralized FL, selecting which clients should train in the next round. For DFL, client selection is particularly effective in non-IID scenarios, where these mechanisms enhance local adaptation while preserving the ability to generalize to unseen data classes \cite{gong2022adaptive}. Among these methods, voting-based training on most selected clients offers an extra on resource usage and performance optimization.

Client selection in DFL presents several challenges related to performance optimization and resource efficiency. One key issue is the trade-off between communication overhead and model quality. While voting mechanisms can improve model convergence by focusing on relevant clients, the frequent exchange of updates and votes among neighbors increases resource consumption, particularly in large-scale networks. Additionally, achieving optimal client selection in highly non-IID settings remains challenging. Clients with rare or unique data distributions may be underrepresented, limiting the model ability to generalize effectively. Furthermore, there is a need to design voting strategies that balance resource usage and local model performance, ensuring efficient training without sacrificing accuracy or diversity. Addressing these challenges is essential to fully realize the potential of voting-based mechanisms in enhancing DFL \cite{fu2023client}.

To improve the previous challenges, this work introduces S-VOTE, a voting-based client selection mechanism to optimize resource usage and improve performance in non-IID scenarios. S-VOTE leverages model similarity for targeted aggregation, ensuring efficient collaboration among relevant clients while reducing unnecessary communication and energy consumption. Additionally, it incorporates an adaptive strategy for spontaneous local training in underutilized clients, addressing participation imbalance without compromising resource efficiency. S-VOTE has been validated through a pool of experiments, demonstrating enhanced convergence, reduced communication and energy overhead, and robust generalization to unseen data. In more detail, S-VOTE achieves up to 21\% improvement in communication efficiency, 4-6\% faster convergence, and 9-17\% higher accuracy in some configurations compared to baseline methods, while also reducing energy consumption by 14-24\%.

The paper is organized as follows. Section~\ref{sec:related} reviews related work on client selection in DFL. Section~\ref{sec:algorithm} details the proposed voting-based client selection mechanism. Section~\ref{sec:validation} presents experimental results, comparing the approach with existing methods. Section~\ref{sec:discussion} discusses the results and limitations, while Section~\ref{sec:conclusion} concludes the paper and suggests future research directions.

\section{Related Work}
\label{sec:related}

This section reviews the literature in DFL and client selection strategies. It highlights existing methods for optimizing client participation in performance and resource usage, focusing on voting-based approaches. \tablename~\ref{tab:comparison} summarizes and compares the main related work.

Several works have reviewed the issue of client selection in FL \cite{fu2023client, gouissem2024comprehensive, li2024comprehensive}. Optimization-based approaches treat the selection process as an optimization task to find the optimal set of clients. However, these methods are slow and impractical in real deployments as they may lead to scalability problems in large scenarios. Importance-based methods prioritize clients according to their influence or relative performance increase. Techniques in this category can use metrics such as reputation or system computational capabilities, but they often require the computation of complex metrics to find good solutions. Clustering-based methods group clients based on similarities in terms of data distribution, model capabilities, or devices. These methods are simpler and may lead to sub-optimal solutions, but are better suited for large scenarios or with many connections among clients, like the DFL case. Finally, other methods use reinforcement learning to find complex solutions for the decision-making process of client selection.

One of the first similarity-based client selection methods was proposed by Sui et al. \cite{sui2022find} and called FedeRiCo. In this approach, clients estimate the utility of other participants using an expectation-maximization approach. However, the number of sampled neighbors was fixed, and resource usage optimization was not considered.

Similar to the voting-based client selection mechanisms presented in this paper, Kharrat et al. \cite{kharrat2024decentralized} proposed generating a collaboration graph for personalized DFL in each client. They used a greedy algorithm for graph construction, computing the reward of using the models from each client. It also includes a client budget to define how many clients are selected in each round. The solution improved performance over FedAvg and other literature aggregation methods. However, they did not focus on resource usage improvement during training as the work proposed in this paper.

Long et al. \cite{long2024decentralized} introduced DA-DPFL, a sparse-to-sparser training scheme for DFL that dynamically reduces the subset of model parameters used during training. The approach focuses on energy efficiency, leveraging dynamic aggregation to retain critical information while minimizing resource usage. Their method demonstrated significant energy savings, achieving up to a 5x reduction compared to DFL baselines while also improving test accuracy. However, the work primarily targets energy efficiency and does not address the challenges of communication overhead or extreme data heterogeneity, which are key aspects tackled in the present study.

Sadiev et al. \cite{sadiev2022decentralized} addressed the challenges of DFL model personalization by introducing a novel penalty that aligns with the structure of the computational network, reducing the communication costs typically incurred by centralized approaches. The work also provided lower bounds on communication and computation costs and proposed provably optimal methods for this problem. While the study offers theoretical insights and efficient solutions for personalized learning, it does not directly focus on improving resource usage or addressing extreme non-IID scenarios, which are central to the contributions of the present work.

Wang et al. \cite{wang2024smart} introduced AFIND+, a sampling and aggregation method for DFL, where helpful neighbors are selected based on their contributions. A greedy selection strategy was implemented to choose clients possessing similar data distributions. However, this work employed local data coresets instead of model similarity and did not consider resource usage optimization.

Despite previous advancements, several limitations remain unaddressed. Current methods often overlook dynamic resource optimization during training, focusing instead on static or narrow criteria. Energy efficiency is frequently prioritized without adequately addressing communication overhead or extreme data heterogeneity. Similarly, theoretical cost optimizations provide valuable insights but often lack practical applicability in diverse, non-IID scenarios. These limitations highlight the need for comprehensive approaches that balance performance and enhance future research.

\section{S-VOTE Design and Functionality}
\label{sec:algorithm}

This section outlines S-VOTE, the proposed mechanism for voting-based client selection and conditional local training in DFL. The approach is designed to optimize resource usage and enhance model performance in federations with non-IID data conditions. The proposed mechanism ensures efficient training while maintaining flexibility and scalability in decentralized environments. Algorithm \ref{algo} describes the complete S-VOTE functionality, which is explained below.

\begin{algorithm}[htpb!]
\caption{S-VOTE: Voting-Based Local Client Selection with Conditional Local Training in DFL}
\begin{algorithmic}[1]
\STATE \textbf{Input:} $T_{\text{init}}, n, t, \tau, V_{\text{min}}, p=0.1$
\STATE \textbf{Output:} Optimized model for each client.

\STATE \textbf{Step 1: Initial Federated Training}
\FOR{$t = 1$ to $T_{\text{init}}$}
    \STATE Perform local updates using Eq.~(2). Share updated models and aggregate received ones using Eq.~(1).
\ENDFOR

\STATE \textbf{Step 2: Local Model Divergence}
\FOR{$t = T_{\text{init}}$ to $T_{\text{init}} + n$}
    \STATE Continue local training with Eq.~(2).
\ENDFOR

\STATE \textbf{Step 3: Model Sharing and Similarity Computation}
\STATE Each client shares its model. Compute similarity using cosine similarity Eq.~(3).

\STATE \textbf{Step 4: Client Selection}
\STATE Select clients based on Eq.~(4) using $\mu$ and $\sigma$ from the distance to received models.

\STATE \textbf{Step 5: Voting and Model Aggregation}
\STATE Aggregate selected models with Eq.~(1). Clients vote for selected peers.

\STATE \textbf{Step 6: Conditional Local Training and Sharing}
\FOR{$t = T_{\text{init}} + n$ to $r$}
    \IF{Votes $\geq V_{\text{min}}$ \textbf{or} neighbors $\leq 2$}
        \STATE Perform local training with Eq.~(2).
    \ELSE
        \STATE Perform random training with probability $p$ (Eq.~(6)).
        \STATE Update $p \gets \min(p + 0.1, 1.0)$ if no training is performed.
    \ENDIF
    \STATE Share updates with all neighbors.
    \STATE Perform aggregation using Eq.~(1). on the models from selected clients.
\ENDFOR
\end{algorithmic}
\label{algo}
\end{algorithm}

\subsection{Initial Federated Training}

In the initial phase of the mechanism, all clients start by training a shared model, which is randomly initialized locally. This stage is essential for all clients to establish a common starting point before local data distributions influence the models. The initial federated training rounds involve aggregating the models of all clients, enabling them to converge toward a base model jointly. This is done by averaging the models received from other clients with the local one. This standard approach in FL is known as \texttt{FedAvg}. The formula for the aggregation step at round $t$ is:

\begin{equation}
\mathbf{w}^{(t)} = \frac{1}{N} \sum_{i=1}^{N} \mathbf{w}_i^{(t-1)},
\end{equation}

where $\mathbf{w}^{(t)}$ represents the global model at round $t$, and $\mathbf{w}_i^{(t-1)}$ is the model of client $i$ at the previous round. $N$ is the total number of neighboring clients. 

After aggregation, the model is broadcast to all clients for further local updates. Note that other aggregation algorithms could be used during this or any other aggregation step.

\subsection{Local Model Divergence}

Once the base model has been established, clients perform local training for a fixed number of rounds. The purpose of local model divergence is to allow each client’s model to adapt to its data distribution, which can differ significantly across clients. This step ensures that the global model does not become overly generalized and retains the unique characteristics of each client’s local data. During this phase, each client updates its model by performing local gradient updates based on its dataset. The local update rule at client $i$ for round $t$ can be expressed as:

\begin{equation}
\mathbf{w}_i^{(t)} = \mathbf{w}_i^{(t-1)} - \eta \nabla L_i(\mathbf{w}_i^{(t-1)}),
\end{equation}

where $\eta$ is the learning rate, and $L_i$ is the loss function for client $i$. After completing the local updates, the client will share the model with its neighbors for further aggregation and selection in subsequent rounds.

\subsection{Model Sharing and Similarity Computation}

After local training, clients share their updated models with their neighbors. This step enables the system to compute the similarity between client models and determine which clients are more similar to each other in terms of their local training. Cosine similarity is commonly used to measure the distance between two models, as it accounts for the directionality of the weight vectors and is invariant to scale. The cosine similarity between two model vectors $\mathbf{w}_i$ and $\mathbf{w}_j$ is defined as:

\begin{equation}
\text{cosine\_similarity}(\mathbf{w}_i, \mathbf{w}_j) = \frac{\mathbf{w}_i \cdot \mathbf{w}_j}{\|\mathbf{w}_i\| \|\mathbf{w}_j\|},
\end{equation}

where $\mathbf{w}_i \cdot \mathbf{w}_j$ is the dot product between the two vectors, and $|\mathbf{w}_i|$ and $|\mathbf{w}_j|$ are the norms of the respective vectors. This measure helps identify clients with similar models based on their local data and training, facilitating efficient client selection in later rounds.

Note that other similarity computation mechanisms could be applied in this step.

\subsection{Client Selection}

The client selection process aims to select the most relevant clients for further processing based on the similarity of their models. The goal is to select clients whose models are sufficiently close to each other to ensure efficient aggregation and reduce unnecessary diversity. The selection criterion is based on the similarity between models, so note that each client has its own set of selected clients. Let $\mu$ be the average cosine similarity of all clients and $\sigma$ the standard deviation. Clients whose similarity exceeds the threshold $\mu + \tau \cdot \sigma$, where $\tau$ is a predefined value, are selected. The selection rule can be written as:

\begin{equation}
\text{Select client } i \text{ if } \text{cosine\_similarity}(\mathbf{w}_l, \mathbf{w}_{\text{i}}) \geq \mu + \tau \cdot \sigma,
\end{equation}

where $\mathbf{w}_{\text{l}}$ is the local model, and $\mu + \tau \cdot \sigma$ represents the upper bound for the similarity threshold. This strategy ensures that only clients whose models are sufficiently close to the local model are selected, facilitating better aggregation and resource efficiency.

\subsection{Voting Mechanism}

Once the neighboring clients are selected, votes are sent to them to indicate their inclusion in the training and aggregation process. The voting mechanism serves as a way to ensure that only clients whose models are sufficiently aligned with the global model will participate in future training. The number of votes a client receives is compared to a vote threshold, $V_{\text{min}}$. If a client receives enough votes, it proceeds with local training. Otherwise, it either engages in random local training or waits for more votes in the following rounds. The voting rule can be expressed as:

\begin{equation}
\text{Client } i \text{ proceeds to local training if } \text{votes}_i \geq V_{\text{min}}.
\end{equation}

This mechanism ensures that only the most relevant clients, based on model similarity, are selected for further training rounds, optimizing the training process.

\subsection{Conditional Local Training}

The conditional local training step determines whether a client should proceed with local training or engage in random training. If a client receives enough votes (i.e., $\text{votes}_i \geq V{\text{min}}$) or has fewer than two neighbors, it will perform local training using the aggregated models of the selected clients. The update rule for local training is the same as in earlier stages, with the model being updated based on the selected clients' aggregated updates. If a client does not meet these conditions, it performs random local training with a probability $p$, which starts at 10\% and increases incrementally by 10\% for each round in which the client does not receive sufficient votes. The update for random training can be expressed as:

\begin{equation}
\text{Local training with } p \quad \text{where} \quad p = \min(p + 0.1, 1.0).
\end{equation}

This randomization helps maintain client participation and avoid stagnation, particularly in cases where a client is not selected due to particular data distributions. Once the local training is completed, the client shares the updated model with its neighbors for further aggregation and collaboration.

\section{S-VOTE Evaluation}
\label{sec:validation}

This section describes the scenario setup and the pool of experiments where the performance of S-VOTE is evaluated. Additionally, it compares the S-VOTE performance to the state of the art (\texttt{FedAvg}, \texttt{SCAFFOLD}, and \texttt{FedProx}). 

\subsection{Validation Scenario}

Evaluations are conducted under non-IID data scenarios, employing the following image-classification datasets: (i) MNIST, comprising 10 classes of $28\times28$ grayscale handwritten digits; (ii) FashionMNIST, containing 10 classes of $28\times28$ grayscale articles of clothing and accessories; (iii) EMNIST, which includes 47 classes of $28\times28$ grayscale handwritten digits and letters; and (iv)~CIFAR10, consisting of 10 classes of $32\times32$ RGB natural images.

Data heterogeneity is introduced by partitioning the datasets among clients using a Dirichlet distribution with a concentration parameter $\alpha \in \{0.1,0.5\}$. Smaller values of $\alpha$ produce more skewed data splits. Each client trains a ResNet9 architecture adapted to the corresponding input shapes and output classes without relying on pre-trained weights. Unless stated otherwise, each client trains for two local epochs per round using a mini-batch size of 32 and a constant learning rate of $10^{-3}$. This ResNet9 includes standard convolutional blocks, batch normalization layers, and skip connections, ensuring sufficient representational capacity for moderate-scale image classification tasks while maintaining computational efficiency.

The experiments span 30 federation rounds, comparing two types of network topologies: (i) fully connected networks comprising 10 and 20 clients, and (ii) random Erd\H{o}s--R\'enyi graphs with 10 and 20 clients using a connection probability of 0.5. A fully connected network of \(N\) clients means that each client maintains direct connections with all others, enabling full information exchange in each round. By contrast, the Erd\H{o}s--R\'enyi setting establishes a subset of connections governed by a probabilistic model, typically reducing communication overhead but introducing synchronization challenges in non-IID scenarios. In the experiments using S-VOTE, the vote threshold $V_{\text{min}}$ is set to balance model similarity-based selection and client participation, ensuring stable training and efficient aggregation. The threshold is determined dynamically based on the number of neighbors $N$, set as $V_{\text{min}} = N/2$ to balance selectivity and inclusivity. This selection strategy prevents excessive client exclusion while ensuring that updates come from a sufficiently diverse subset of well-aligned models. Empirical results confirm that this threshold optimally stabilizes training, accelerates convergence, and enhances model performance across varying degrees of non-IID data distributions.

The experiments are conducted on a computational cluster equipped with Intel i7-10700F CPU, two RTX 3080 GPUs, and 98GB RAM to ensure consistent computational and communication dynamics. All experiments are implemented within the Nebula framework, which supports decentralized client orchestration and real network communications among containerized clients \cite{beltran2024fedstellar}. 

Additionally, the framework integrates an energy consumption monitoring module, measuring power usage in kilowatt-hours (kWh) to evaluate DFL efficiency \cite{Chen24}. Total energy consumption is assessed across three phases: (i) local training, which estimates energy using GPU performance counters and model, capturing power draw during computations; (ii) aggregation, which accounts for energy used in model averaging and weight updates across clients; (iii) communication, which measures energy spent on transmitting updates, influenced by the number of neighbors, federation rounds, total bytes exchanged, and per-byte transmission cost. The primary comparison focuses on model performance under varying non-IID settings, convergence speed, and total communication overhead during federation.

\subsection{Results and Discussion}
\label{sec:discussion}

This section analyzes the results obtained from the experiments, highlighting the strengths and limitations of the proposed voting-based client selection mechanism. It discusses the model performance, communication costs, and convergence speed while highlighting the improvements and addressing potential challenges for further refinement.

\subsubsection{Model performance}

\begin{table*}[h]
    \centering
    \scriptsize
    \renewcommand{\arraystretch}{1.2}
    \setlength{\tabcolsep}{3pt}
    \caption{Model Performance (F1 score) for Different Data Distributions, Datasets, Number of Clients, and Methods.}
    \label{tab:exp_results}
    \resizebox{\textwidth}{!}{
    \begin{threeparttable}
    \begin{tabular}{cc|cccc|cccc|cccc|cccc}
        \toprule
        \multicolumn{2}{c|}{\textbf{Data Distribution}} & \multicolumn{16}{c}{\textbf{Dirichlet $\alpha = 0.5$}}  \\
        \midrule
        \multicolumn{2}{c|}{\textbf{Dataset}} & \multicolumn{4}{c|}{MNIST} & \multicolumn{4}{c|}{FashionMNIST} & \multicolumn{4}{c|}{EMNIST} & \multicolumn{4}{c}{CIFAR-10} \\
        \midrule
        \multicolumn{2}{c|}{\textbf{Topology@Clients}} & F@10 & F@20 & R@10 & R@20 & F@10 & F@20 & R@10 & R@20 & F@10 & F@20 & R@10 & R@20 & F@10 & F@20 & R@10 & R@20 \\
        \midrule
        \multirow{3}{*}{\rotatebox[origin=c]{90}{\textbf{Method}}} 
        & FedAvg & 0.98$\pm$0.21 & 0.97$\pm$0.22 & 0.89$\pm$0.17 & 0.85$\pm$0.22 & 0.91$\pm$0.16 & 0.90$\pm$0.18 & 0.82$\pm$0.18 & 0.85$\pm$0.17 & 0.76$\pm$0.19 & 0.68$\pm$0.20 & 0.66$\pm$0.17 & 0.56$\pm$0.21 & 0.69$\pm$0.18 & 0.65$\pm$0.17 & 0.63$\pm$0.16 & \textbf{0.63$\pm$0.14} \\
        & SCAFFOLD & 0.98$\pm$0.29 & 0.98$\pm$0.34 & 0.97$\pm$0.30 & \textbf{0.98$\pm$0.34} & 0.92$\pm$0.23 & \textbf{0.92$\pm$0.26} & 0.90$\pm$0.23 & \textbf{0.91$\pm$0.28} & \textbf{0.85$\pm$0.25} & \textbf{0.84$\pm$0.25} & 0.73$\pm$0.29 & 0.74$\pm$0.26 & \textbf{0.75$\pm$0.21} & \textbf{0.70$\pm$0.21} & \textbf{0.71$\pm$0.23} & \textbf{0.63$\pm$0.19} \\
        & FedProx & \textbf{0.99$\pm$0.29} & 0.97$\pm$0.35 & \textbf{0.98$\pm$0.33} & \textbf{0.98$\pm$0.36} & 0.92$\pm$0.23 & \textbf{0.92$\pm$0.28 }& \textbf{0.91$\pm$0.27} & \textbf{0.91$\pm$0.27} & 0.83$\pm$0.24 & 0.82$\pm$0.17 & 0.73$\pm$0.27 & 0.73$\pm$0.27 & 0.66$\pm$0.22 & 0.51$\pm$0.18 & 0.43$\pm$0.17 & 0.43$\pm$0.17 \\
        \rowcolor{Gray} & S-VOTE & \textbf{0.99$\pm$0.21} & \textbf{0.99$\pm$0.22 }& \textbf{0.98$\pm$0.18} & 0.93$\pm$0.22 & \textbf{0.98$\pm$0.17} & \textbf{0.92$\pm$0.17} & 0.88$\pm$0.18 & \textbf{0.91$\pm$0.17} & 0.83$\pm$0.21 & 0.77$\pm$0.21 & \textbf{0.74$\pm$0.21} & \textbf{0.75$\pm$0.21} & \textbf{0.75$\pm$0.18} & 0.65$\pm$0.18 & 0.64$\pm$0.14 & \textbf{0.63$\pm$0.12} \\
        \bottomrule
        \toprule
        \multicolumn{2}{c|}{\textbf{Data Distribution}} & \multicolumn{16}{c}{\textbf{Dirichlet $\alpha = 0.1$}}  \\
        \midrule
        \multicolumn{2}{c|}{\textbf{Dataset}} & \multicolumn{4}{c|}{MNIST} & \multicolumn{4}{c|}{FashionMNIST} & \multicolumn{4}{c|}{EMNIST} & \multicolumn{4}{c}{CIFAR-10} \\
        \midrule
        \multicolumn{2}{c|}{\textbf{Topology@Clients}} & F@10 & F@20 & R@10 & R@20 & F@10 & F@20 & R@10 & R@20 & F@10 & F@20 & R@10 & R@20 & F@10 & F@20 & R@10 & R@20 \\
        \midrule
        \multirow{3}{*}{\rotatebox[origin=c]{90}{\textbf{Method}}} 
        & FedAvg & 0.79$\pm$0.22 & 0.87$\pm$0.25 & 0.63$\pm$0.19 & 0.68$\pm$0.19 & 0.57$\pm$0.16 & 0.75$\pm$0.17 & 0.44$\pm$0.07 & 0.49$\pm$0.16 & 0.57$\pm$0.16 & 0.67$\pm$0.19 & 0.42$\pm$0.09 & 0.49$\pm$0.18 & 0.29$\pm$0.08 & 0.36$\pm$0.09 & 0.27$\pm$0.08 & 0.30$\pm$0.03 \\
        & SCAFFOLD & 0.81$\pm$0.34 & 0.84$\pm$0.38 & 0.82$\pm$0.34 & 0.65$\pm$0.37 & 0.84$\pm$0.25 & 0.86$\pm$0.28 & 0.69$\pm$0.25 & 0.58$\pm$0.27 & 0.75$\pm$0.22 & \textbf{0.77$\pm$0.26} & 0.58$\pm$0.26 & 0.59$\pm$0.24 & \textbf{0.40$\pm$0.13} & \textbf{0.42$\pm$0.12} & \textbf{0.36$\pm$0.17} & 0.33$\pm$0.07 \\
        & FedProx & \textbf{0.97$\pm$0.33} & \textbf{0.98$\pm$0.33} & \textbf{0.97$\pm$0.34 }& \textbf{0.97$\pm$0.39} & \textbf{0.88$\pm$0.26} & 0.79$\pm$0.30 & 0.79$\pm$0.22 & 0.67$\pm$0.25 & \textbf{0.75$\pm$0.24} & 0.75$\pm$0.25 & 0.59$\pm$0.26 & 0.64$\pm$0.26 & 0.27$\pm$0.10 & 0.09$\pm$0.04 & 0.15$\pm$0.07 & 0.10$\pm$0.03 \\
        \rowcolor{Gray} & S-VOTE & 0.94$\pm$0.21 & 0.89$\pm$0.23 & 0.93$\pm$0.19 & 0.86$\pm$0.19 & 0.84$\pm$0.13 & \textbf{0.88$\pm$0.17} & \textbf{0.80$\pm$0.07} & \textbf{0.84$\pm$0.16} & 0.65$\pm$0.16 & 0.67$\pm$0.19 & \textbf{0.65$\pm$0.09} & \textbf{0.66$\pm$0.18 }& 0.31$\pm$0.08 & 0.36$\pm$0.08 & 0.33$\pm$0.08 & \textbf{0.38$\pm$0.03} \\
        \bottomrule
    \end{tabular}
    \begin{tablenotes}
        \item R: Random, F: Fully-connected
    \end{tablenotes}
    \end{threeparttable}}
\end{table*}

\begin{figure*}[t!]
    \centering
    \includegraphics[width=0.97\textwidth]{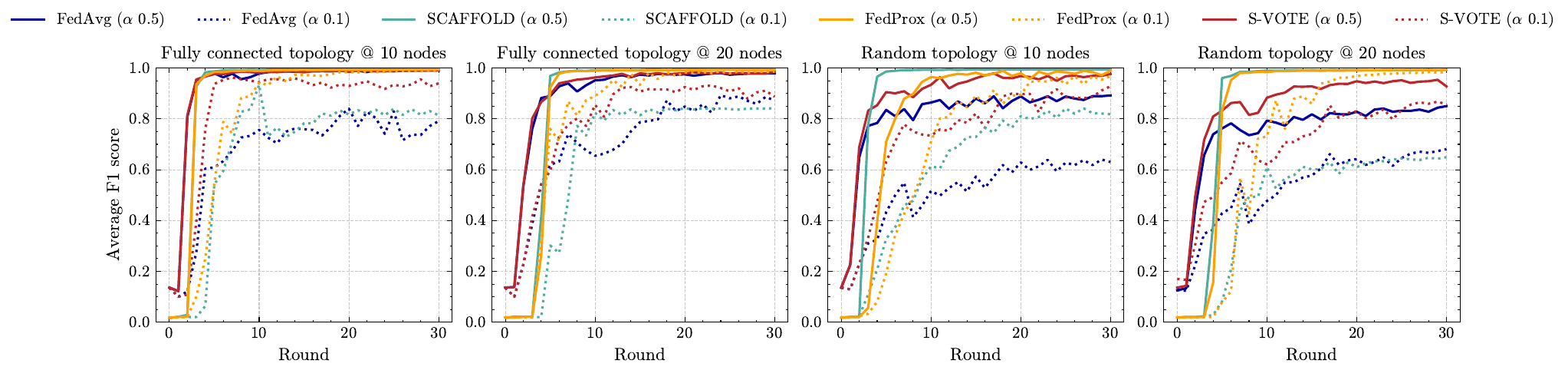}
    \caption{F1-Score Comparison Between the Proposed Mechanism (S-VOTE) and SoTA Aggregation Algorithms using MNIST}
    \label{fig:validation-MNIST}
\end{figure*}

\begin{figure*}[t!]
    \centering
    \includegraphics[width=0.97\textwidth]{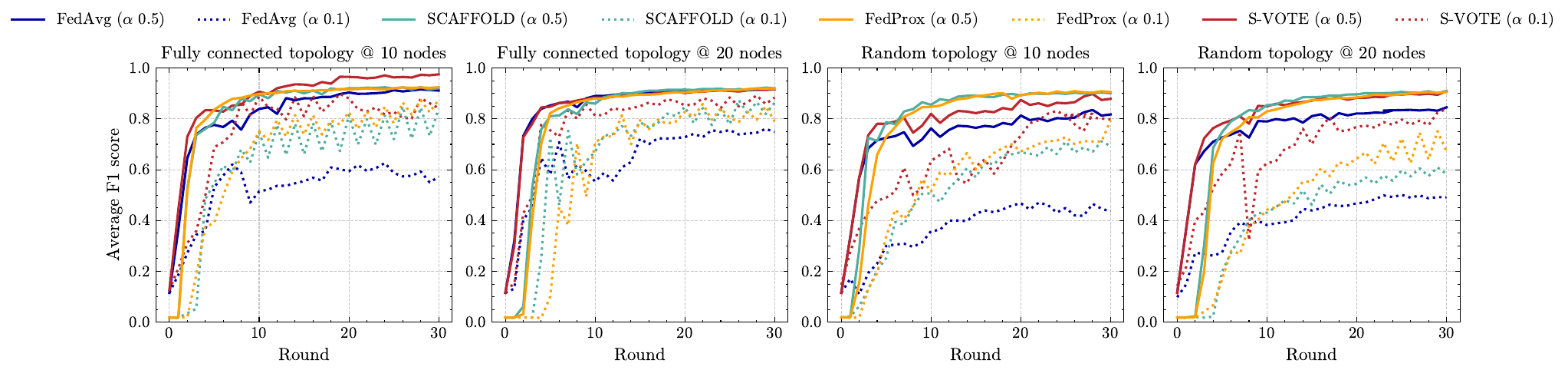}
    \caption{F1-Score Comparison Between the Proposed Mechanism (S-VOTE) and SoTA Aggregation Algorithms using FashionMNIST}
    \label{fig:validation-FashionMNIST}
\end{figure*}

\tablename~\ref{tab:exp_results} summarizes all the results achieved by the different aggregation methods in the experiments performed. S-VOTE consistently achieves faster convergence across all configurations using MNIST (see \figurename~\ref{fig:validation-MNIST}). In the fully connected topology with 10 clients, S-VOTE reaches an F1-score of 99\% within the first rounds for $\alpha = 0.5$, similar to FedAvg with 98\%. The difference becomes more evident for $\alpha = 0.1$, where FedAvg stabilizes at 79\%, while S-VOTE achieves 94\%. In the 20-client fully connected topology, S-VOTE maintains its superior performance, converging at 99\%, while FedAvg remains at 97\%, SCAFFOLD at 98\%, and FedProx is slightly behind. The advantage of S-VOTE becomes apparent in the early rounds. This suggests that similarity-based voting facilitates faster alignment between participating clients, reducing the impact of heterogeneous data partitions. The trends persist in the FashionMNIST dataset (see \figurename~\ref{fig:validation-FashionMNIST}). In the fully connected topology with 10 clients, S-VOTE reaches an F1-score of 98\% for $\alpha = 0.5$ and 84\% for $\alpha = 0.1$, surpassing FedAvg at 91\% and 57\%, respectively. In the fully connected topology with 20 clients, S-VOTE reaches 91\% and 88\%, outperforming FedAvg and getting the same values as SCAFFOLD and FedProx. In random topologies, the performance advantage of S-VOTE becomes even more evident when $\alpha = 0.1$, confirming that model similarity-based selection stabilizes learning in decentralized environments with non-IID data. 

\begin{figure*}[t!]
    \centering
    \includegraphics[width=0.97\textwidth]{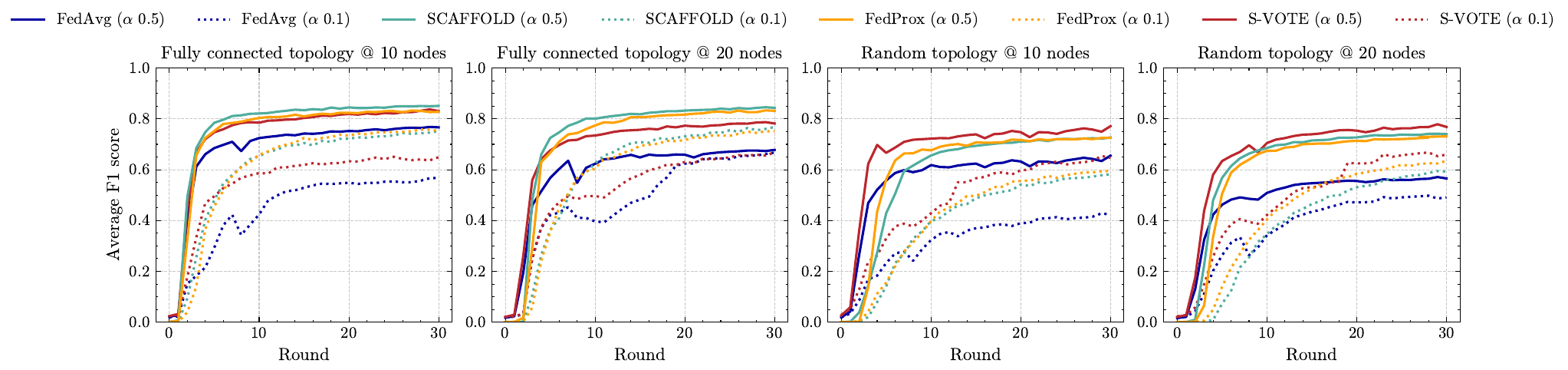}
    \caption{F1-Score Comparison Between the Proposed Mechanism (S-VOTE) and SoTA Aggregation Algorithms using EMNIST}
    \label{fig:validation-EMNIST}
\end{figure*}

\begin{figure*}[t!]
    \centering
    \includegraphics[width=0.97\textwidth]{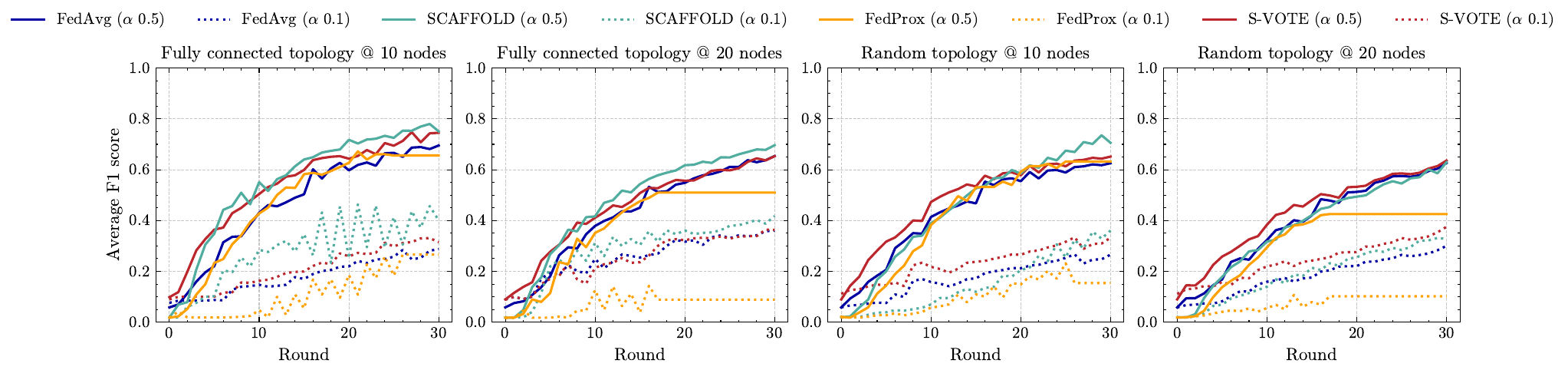}
    \caption{F1-Score Comparison Between the Proposed Mechanism (S-VOTE) and SoTA Aggregation Algorithms using CIFAR10}
    \label{fig:validation-CIFAR10}
\end{figure*}

For the EMNIST dataset (see \figurename~\ref{fig:validation-EMNIST}), the relative advantages of S-VOTE decrease compared to other algorithms. In fully connected topologies, S-VOTE converges at 83\% for $\alpha = 0.5$ and 10 clients, exceeding FedAvg at 76\%. In this case, SCAFFOLD overpasses S-VOTE, while SCAFFOLD and FedProx obtain an increase of 9\% with 20 clients. However, it is noteworthy that in the early training rounds, FedProx occasionally outperforms S-VOTE, suggesting that its regularization term provides short-term stabilization in extreme non-IID conditions. Finally, S-VOTE demonstrates performance stability in all the experiments using the CIFAR10 dataset (see \figurename~\ref{fig:validation-CIFAR10}). Despite this, the SCAFFOLD algorithm overpasses S-VOTE around 5\%. In the fully connected topology with 10 clients, S-VOTE achieves an F1-score of 75\% for $\alpha = 0.5$, outperforming FedAvg at 69\%, though FedProx reaches 66\%, making it the worst-performing method. In the random topology with 20 clients, S-VOTE stabilizes at 63\%, similar to FedAvg and SCAFFOLD, exceeding FedProx performance with 43\%.

\subsubsection{Convergence Time, Energy, and Communications}

The comparison of elapsed time and energy consumption focuses on FedAvg and S-VOTE, as FedAvg represents an upper bound on training and communication time (see \tablename~\ref{tab:elapsed_time}). Other methods, such as SCAFFOLD and FedProx, introduce additional computational overhead—SCAFFOLD requires transmitting a control variate parameter, increasing network costs. At the same time, FedProx adds a regularization term, affecting local training time. Since these modifications inherently raise resource demands, FedAvg serves as the most relevant baseline for assessing the efficiency of S-VOTE.

The results show that S-VOTE reduces training time across all datasets while significantly lowering energy consumption. For MNIST, S-VOTE achieves a 5.19\% reduction in elapsed time, decreasing from 1691.3s to 1603.5s, with a substantial 21.30\% drop in energy consumption. A similar trend is observed for FashionMNIST, where training time is reduced by 4.16\%, and energy consumption decreases by 20.96\%, confirming that selective client participation enhances efficiency without introducing unnecessary computational overhead. 

In EMNIST, S-VOTE leads to a 5.72\% reduction in elapsed time while significantly lowering energy consumption by 23.57\%. This improvement suggests that despite the increased complexity of character recognition tasks, S-VOTE efficiently balances client selection and resource utilization, making it more energy-efficient while maintaining competitive convergence times. Finally, in CIFAR-10, where computational demands are higher due to RGB image processing, S-VOTE reduces training time by 6.17\% while lowering energy consumption by 14.37\%. The reduction in elapsed time across all datasets can be attributed to the fact that clients do not wait for updates from all their neighbors, only from the selected clients based on the voting mechanism. This optimization minimizes idle time and reduces unnecessary computations, leading to faster convergence without sacrificing model performance.

\begin{table}[htpb]
    \centering
    \scriptsize
    \caption{Average Elapsed Time (s) and Energy Consumption (kWh) under Non-IID Distribution ($\alpha\; 0.5$) compared with FedAvg}
    \label{tab:elapsed_time}
    \resizebox{\columnwidth}{!}{
    \begin{threeparttable}
    \begin{tabular}{@{}lcccc@{}}
        \toprule
        \multirow{2}{*}{} & \multicolumn{2}{c}{FedAvg} & \multicolumn{2}{c}{\textbf{S-VOTE}} \\
        \cmidrule(lr){2-3} \cmidrule(lr){4-5}
        & Time & Energy\textsuperscript{$\dagger$} & Time & Energy\textsuperscript{$\dagger$} \\
        \midrule
        
        MNIST & 1691.3$\pm$24.1 & 0.2725 & 1603.5$\pm$11.4 {\color{teal}($\downarrow$ 5.19\%)} & 0.2145 {\color{teal}($\downarrow$ 21.30\%)} \\
        
        FashionMNIST & 1721.2$\pm$12.6 & 0.2916 & 1649.6$\pm$47.1 {\color{teal}($\downarrow$ 4.16\%)} & 0.2305 {\color{teal}($\downarrow$ 20.96\%)} \\
        
        EMNIST & 2879.8$\pm$17.3 & 0.3941 & 2715.2$\pm$22.0 {\color{teal}($\downarrow$ 5.72\%)} & 0.3012 {\color{teal}($\downarrow$ 23.57\%)} \\
        
        CIFAR10 & 3514.7$\pm$11.9 & 0.5163 & 3297.8$\pm$11.23 {\color{teal}($\downarrow$ 6.17\%)} & 0.4421 {\color{teal}($\downarrow$ 14.37\%)} \\
        
        \bottomrule
    \end{tabular}
    \begin{tablenotes}
        \item \textsuperscript{$\dagger$} Total sum of energy consumption of all clients in the federation.
    \end{tablenotes}
    \end{threeparttable}}
\end{table}

\tablename~\ref{tab:network} shows the evaluation of network communication costs focusing on the total bytes received and sent per client under FedAvg and S-VOTE. Since FedAvg uniformly aggregates updates from all clients, it represents an upper bound on communication overhead, making it a suitable baseline for comparison.

\begin{table}[htpb]
    \centering
    \scriptsize
    \caption{Average Network Bytes Received and Sent (GB) under Non-IID Distribution ($\alpha\; 0.5$) compared with FedAvg}
    \resizebox{\columnwidth}{!}{
    \begin{tabular}{@{}lcccc@{}}
        \toprule
        \multirow{2}{*}{} & \multicolumn{2}{c}{FedAvg} & \multicolumn{2}{c}{\textbf{S-VOTE}} \\
        \cmidrule(lr){2-3} \cmidrule(lr){4-5}
        & Bytes Received & Bytes Sent & Bytes Received & Bytes Sent \\
        \midrule
        MNIST & 3.62$\pm$0.2 & 3.65$\pm$0.1 & 3.05$\pm$0.2 {\color{teal}($\downarrow$ 15.75\%)} & 3.09$\pm$0.2 {\color{teal}($\downarrow$ 15.34\%)} \\
        FashionMNIST & 3.67$\pm$0.2 & 3.65$\pm$0.1 & 3.13$\pm$0.1 {\color{teal}($\downarrow$ 14.71\%)} & 3.22$\pm$0.1 {\color{teal}($\downarrow$ 11.78\%)} \\
        EMNIST & 3.75$\pm$0.1 & 3.71$\pm$0.2 & 3.05$\pm$0.2 {\color{teal}($\downarrow$ 18.67\%)} & 3.09$\pm$0.1 {\color{teal}($\downarrow$ 16.71\%)} \\
        CIFAR10 & 3.91$\pm$0.2 & 3.87$\pm$0.1 & 3.12$\pm$0.2 {\color{teal}($\downarrow$ 20.20\%)} & 3.07$\pm$0.2 {\color{teal}($\downarrow$ 20.67\%)} \\
        \bottomrule
    \end{tabular}}
    \label{tab:network}
\end{table}

The results confirm that S-VOTE significantly reduces communication overhead across all datasets by optimizing client selection and minimizing unnecessary transmissions. For MNIST, bytes received and sent decrease by 15.75\% and 15.34\%, respectively, while FashionMNIST sees reductions of 14.71\% and 11.78\%, demonstrating that the voting mechanism effectively balances communication efficiency and model updates. In EMNIST, where task complexity increases due to a larger number of classes, bytes received drop by 18.67\% and bytes sent by 16.71\%, highlighting that S-VOTE efficiently reduces redundant updates in high-class diversity tasks. The most significant reductions occur in CIFAR-10, with a 20.20\% decrease in bytes received and 20.67\% in bytes sent, reinforcing that selective voting minimizes bandwidth consumption even in high-dimensional tasks.

\section{Conclusion and Future Work}
\label{sec:conclusion}

This work introduced S-VOTE, a voting-based client selection mechanism for DFL, designed to mitigate the challenges posed by non-IID data distributions, communication overhead, and resource constraints. By leveraging model similarity-based voting, S-VOTE optimizes client participation, reducing unnecessary updates and improving convergence efficiency. Additionally, its adaptive local training strategy ensures a balanced contribution from underutilized clients without imposing significant resource costs. Extensive experiments on benchmark datasets, including MNIST, FashionMNIST, EMNIST, and CIFAR-10, demonstrate that S-VOTE achieves up to 21\% lower communication costs, 4-6\% faster convergence, and 9-17\% improved accuracy compared to baseline methods in some configurations, while also reducing energy consumption by up to 24\%. These results confirm that S-VOTE enhances model generalization and training stability, making it a strong candidate for practical decentralized learning in heterogeneous and resource-constrained environments.

While S-VOTE effectively reduces communication and improves convergence, further optimizations can enhance efficiency. Future work will focus on incorporating dynamic non-IID assessment mechanisms that determine the degree of heterogeneity in each client data at the beginning of the federation. This information can be leveraged to form similarity-based client clusters, optimizing training efficiency by allowing more structured and adaptive client participation. Another potential extension is the integration of adaptive thresholding techniques, where the vote threshold is dynamically adjusted based on real-time model divergence, ensuring balanced participation while maintaining computational efficiency.


\bibliographystyle{IEEEtran}  
\bibliography{references}

\end{document}